%% file: main.tex
\documentclass[letterpaper, 10 pt, conference]{IEEEtran}  

\IEEEoverridecommandlockouts                              

\usepackage{graphics} 
\usepackage{epsfig} 
\usepackage{mathptmx} 
\usepackage{times} 
\usepackage{amsmath} 
\usepackage{amssymb}  

\usepackage[T1]{fontenc}
\usepackage{amsfonts}
\usepackage{booktabs}
\usepackage{siunitx}
\usepackage{caption} 
\usepackage{booktabs}
\usepackage{graphicx}
\usepackage{float}
\usepackage[percent]{overpic}
\usepackage{caption}
\usepackage[font=small,skip=0pt]{subcaption}

\usepackage{array, tabularx, makecell, booktabs}%

\usepackage{geometry}
\usepackage{comment}

\usepackage[table,xcdraw]{xcolor}
\usepackage{multirow}
\usepackage[ruled,noresetcount]{algorithm2e}
\usepackage{algpseudocode}
\algdef{SE}[DOWHILE]{Do}{doWhile}{\algorithmicdo}[1]{\algorithmicwhile\ #1}
\usepackage{hyperref}

\title{\LARGE \bf
Deep-Reinforcement-Learning-based Path Planning for Industrial Robots using Distance Sensors as Observation
}

\author{Teham Bhuiyan$^{1*}$\thanks{$^{1}$ Linh K{\"a}stner, Yifan Hu, Benno Kutschank and Jens Lambrecht are with the Chair Industry Grade Networks and Clouds, Faculty of Electrical Engineering, and Computer Science,				
		Berlin Institute of Technology, Berlin, Germany
		{\tt\small teham@tu-berlin.de}}, Linh K{\"a}stner$^{1}$, Yifan Hu$^{1}$, Benno Kutschank$^{1}$ and Jens Lambrecht$^{1}$
}

\begin{document}

\maketitle
\thispagestyle{empty}
\pagestyle{empty}

\input{content/0-abstract}
\input{content/1-introduction}
\input{content/2-Related-Works}
\input{content/3-methodology}
\input{content/4-evaluations}

\input{content/5-conclusion}


\addtolength{\textheight}{-1cm} 




\typeout{}
\bibliographystyle{IEEEtran}
\bibliography{main}

\end{document}

%% file: content/0-abstract.tex

\begin{abstract}
Industrial robots are widely used in various manufacturing environments due to their efficiency in doing repetitive tasks such as assembly or welding. A common problem for these applications is to reach a destination without colliding with obstacles or other robot arms. Most commonly used sampling-based path planning approaches such as RRT require long computation times, especially in complex environments. Furthermore, the environment in which they are employed needs to be known beforehand. When utilizing the approaches in new environments, a tedious engineering effort in setting hyperparameters needs to be conducted, which is time- and cost-intensive. On the other hand, Deep Reinforcement Learning has shown remarkable results in dealing with unknown environments, generalizing new problem instances, and solving motion planning problems efficiently. On that account, this paper proposes a Deep-Reinforcement-Learning-based motion planner for robotic manipulators. We evaluated our model against state-of-the-art sampling-based planners in several experiments. The results show the superiority of our planner in terms of path length and execution time.

\end{abstract}

%% file: content/1-introduction.tex


\section{Introduction}
\noindent Industrial robots are widely used in manufacturing processes. While robots can mostly perform specific tasks automatically, it is necessary to program them manually. Whereas sampling-based planning approaches can cope well with simple environments, reliable motion planning in complex and unknown scenarios remain a big challenge due to their computational complexity and unexpected dynamics. In particular, for every new environment, industrial robots need to be reprogrammed using methods such as offline programming. These methods require skilled workers to set up a variety of hyperparameters manually, which constitutes a tedious engineering effort and is both time- and cost-consuming. Furthermore, classic approaches such as RRT can only be employed in known environments. On the other hand, Deep Reinforcement Learning (DRL) emerged as an end-to-end learning approach, capable of directly mapping sensor data to robot actions, and has shown promising results in teaching complex behavior rules, increasing robustness to noise, and generalizing new problem instances \cite{everett2018motion}, \cite{chiang2019learning}, \cite{kastner2021arena}. In this paper, we propose a DRL-based approach to automate the motion planning of robot arms. 
\begin{figure}[ht]
	\centering
	\includegraphics[width=0.5\textwidth]{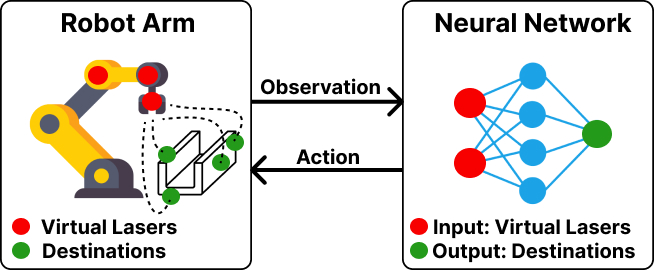}
	\caption{This work proposes DRL-based motion planning using virtual lasers as observation. Given a goal and the virtual sensor observations, the trained agent is able to provide a collision-free path toward the goal.}
	\label{intro}
\end{figure}
In particular, we introduce a training pipeline consisting of virtual laser observations to make the training process more efficient and evaluate the approach on a realistic 3D simulation in various industry-oriented tasks. 

The main contributions of this work are the following:
\begin{itemize}
    \item Realization of realistic industrial robot tasks within the 3D simulator Pybullet
    \item Proposal of an efficient 3D motion planning algorithm using laser scans as observation
    \item Extensive evaluation of the approach against classic baseline approaches in complex scenarios
\end{itemize}

\noindent The paper is structured as follows. Sec. 2 begins with related works followed by the methodology in Sec. 3. Subsequently, the results and evaluations are presented in Sec 4. Finally, Sec. 6 provides a conclusion and outlook.

%% file: content/2-Related-Works.tex
\section{Related Works}
\noindent Collision-free path planning algorithms for robot manipulators have been extensively studied in various research publications. Conventional methods like the artificial potential field and sampling-based algorithms are among the most commonly used in currently employed industrial robot arms\cite{rrt1} \cite{rrt2}\cite{rrt5}.
Recent publications include the work of Xinyu et al. \cite{rrt6}.  The researchers propose a randomization algorithm, P-RT*-connect based on RRT and APF. The motion path is found by exploring two path trees from the start node and destination node with
RRT*. Another recent optimization for RRT planners is proposed by Wang et al. \cite{rrt7b}. The approach is a modified version of the Bidirectional RRT algorithm (Bi-RRT, also known as RRT-ConCon), which modifies the standard RRT algorithm by adding another RRT that grows from the goal configuration with
both trees being expanded at the same rate. Salehian et al. \cite{rrt3} propose a unified framework for coordinated
multi-arm motion planning by providing a centralized inverse kinematic solver under self-collision avoidance constraints in real time.
Zhang et al. \cite{rrt7} propose a real-time kinematic
control strategy for robot manipulators in a dynamic environment.
While sampling-based methods are capable of collision-free planning, they remain slow and inefficient, especially in complex environments. Most sampling-based algorithms assume complete knowledge of the environment, which may not be given in certain scenarios. DRL has emerged as an end-to-end approach with the potential to learn complex behavior in unknown environments. 
In the field of robot navigation, DRL-based approaches have shown promising results  \cite{kastner2022arena}, \cite{dugas2021navrep}, \cite{everett2018motion}, \cite{kastner2022all}.
Wen et al. propose the use of DDPG to plan the trajectory of a robot arm to realize obstacle avoidance \cite{drl5} based purely on DRL.
Bianca Sangiovanni et al. \cite{drl11} propose a hybrid control
methodology to achieve full-body collision avoidance
in anthropomorphic robot manipulators. They combine classical motion planning algorithms with
a Deep Reinforcement Learning (DRL) approach
 to perform obstacle avoidance, while
achieving a reaching task in the operative space. Similar works by Faust et al. \cite{faust2018prm} and Chiang et al. \cite{chiang2019learning} combine DRL-based motion planning with classic approaches such as RRT and PRM for the motion planning of ground robots over long distances. More recently, works by Kästner et al. \cite{kastner2022arena}, Dugas et al. \cite{dugas2020navrep}, and Guldenring et al. \cite{guldenringlearning} showed the superiority of DRL approaches for fast obstacle avoidance in unknown and dynamic environments. DRL-based approaches have also been utilized in a number of research works for motion planning and collision avoidance for stationary robots.
Kamali and Bonev \cite{drl9} propose a DRL-based method to solve the problem of robot
arm motion planning in telemanipulation applications. They map human hand motions to a robot arm in real-time, while avoiding collisions. A drawback of this approach is that obstacles need to be static and predefined. 
Zhong et al. \cite{drl14} propose a path planner for welding manipulators based on DRL. They demonstrated in three specific welding tasks that DRL-based models can achieve competitive results compared to sampling-based approaches.  Meyesa et al. \cite{drl13} present an approach based on RL and Q-learning enabling an agent to
control a six-axis industrial robot to play the wire loop game using a camera.   
\noindent Zhu et al. \cite{drl16} propose a novel collision avoidance framework using a depth camera that
allows robots to work alongside human operators in unstructured and complex environments. Of note, the above works rely on sensors to perceive the environment and act accordingly. However, for industrial tasks such as welding, the use of static sensors is sub-optimal as the robot may move outside of the sensor's range or sensors may not be able to detect areas in narrow spaces. For this reason, we propose an approach where virtual laser sensors are strategically placed on different parts of the robot arm. This way the sensors move alongside the arm enabling the robot to perform collision-free motion planning in complex environments.

%% file: content/3-methodology.tex
\begin{figure*}[!htp]
	\centering
	\includegraphics[width=0.96\textwidth]{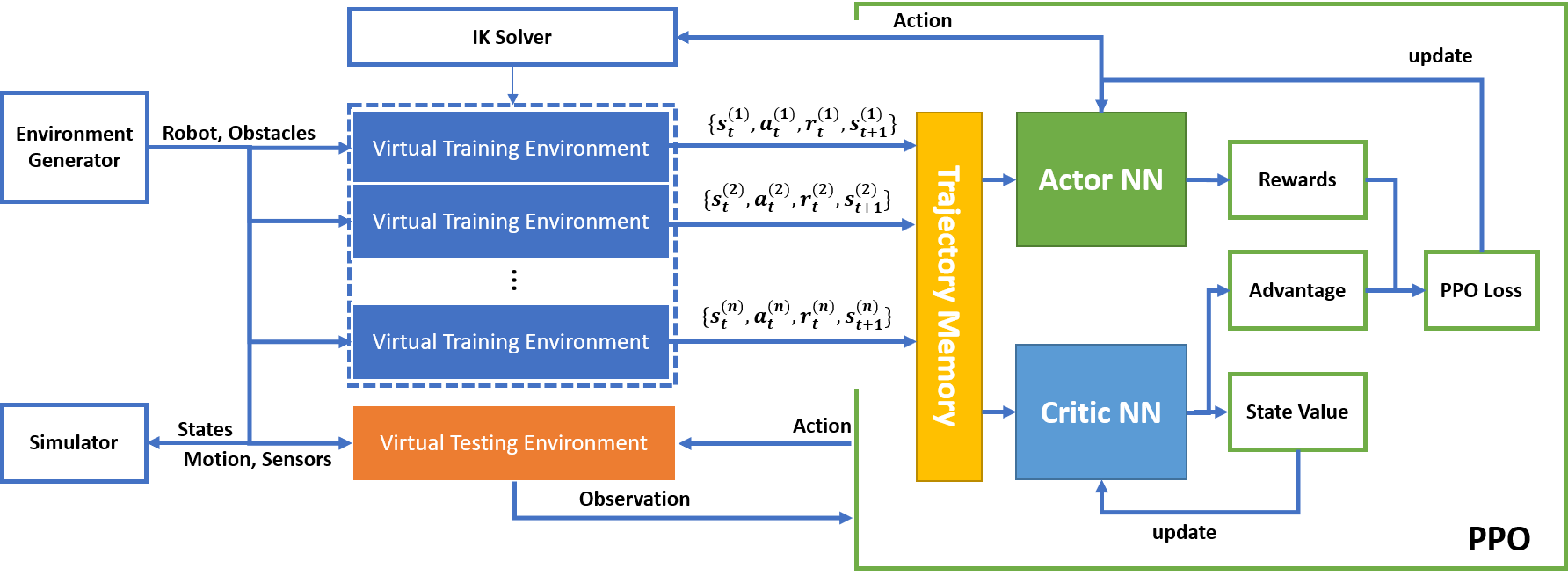}
	\caption{System design of our proposed approach. For observations, we use three virtual laser sensors (two on the end effector and one on top of the wrist, which is visualized in Fig. \ref{fig:vlsr}). We employ an actor-critic approach where the agent outputs continuous values predicting the change of the robot's current pose. The motion of the robot is realized by the inverse kinematics (IK) solver. This process is repeated until the robot reaches its goal.}
	\label{system}
\end{figure*}

\section{Methodology}
\noindent In this chapter, the methodology of the proposed framework is presented. The system design is described in Fig. \ref{system}. We train a DRL agent to perform collision-free motion planning for industrial robots only using virtual laser scan observations. Training and testing are done in the simulator Bullet3 \cite{pb}. 

\subsection{Robot Kinematics}
\noindent Robot control algorithms rely on kinematics to calculate accurate commands. A common way to model these for robots composed out of solid links and joints is the Denavit-Hartenberg (D-H) representation. It uses homogeneous transformation matrices, which are functions of the robot configuration $\mathbf{q} = [q_1, q_2, \dots, q_n]$, to describe the relationship between the coordinate systems of two adjacent links. Chain-multiplying these matrices for a given robot yields the transformation from the robot's base link to its end effector, which is the control target in most applications. For a 6-degree-of-freedom robot that transformation from base to end effector is given by
\begin{align}
    ^{0}_{\textrm{E}}\mathbf{T}(\mathbf{q}) = (\prod^{6}_{i=1} \ ^{i-1}_i\mathbf{T}(q_i)) \cdot ^{6}_{\textrm{E}}\mathbf{T}, \label{fk}
\end{align}

with a single matrix $^{i-1}_i\mathbf{T}(q_i)$ defined by

\begin{align}
    ^{i-1}_i\mathbf{T}(q_i) = 
    \begin{bmatrix}
        \cos(\theta_i) & \sin(\theta_i) & 0 & 0 \\
        -\sin(\theta_i) \cos(\alpha_i) & \cos(\theta_i) \cos(\alpha_i) & \sin(\alpha_i) & 0\\
        \sin(\theta_i) \sin(\alpha_i) & \sin(\alpha_i) & \cos(\alpha_i) & 0 \\
        a_i \cos(\theta_i) & a_i \sin(\theta_i) & d_i & 1
    \end{bmatrix}^t.
\end{align}

\noindent Equation (\ref{fk}) can be understood as the forward kinematics of a robot: for a given configuration $\mathbf{q}$, the transformation $^{0}_{\textrm{E}}\mathbf{T}(\mathbf{q})$ gives the orientation (the upper 3x3 partial matrix) and the position (the first three elements of the rightmost column) of the end effector in the robot's base frame.
However, most tasks in robotics are defined in the workspace, not the configuration space. This necessitates inverse kinematics, such that trajectories defined in a workspace can be mapped into configuration space and subsequently used for control inputs. Generally speaking, in inverse kinematics a desired workspace displacement $\dot{\mathbf{x}}$ is related to an at first unknown configuration space displacement $\dot{\mathbf{q}}$ by
\begin{align}
    \dot{\mathbf{x}} = \mathbf{J}(\mathbf{q}) \cdot \dot{\mathbf{q}}, \label{jac}
\end{align}
where $\mathbf{J}(\mathbf{q})$ is the robot's Jacobian. As exact solutions are often impractical, multiple ways to numerically solve this equation for $\dot{\mathbf{q}}$ have been found. The method chosen for this work is the Damped Least Squares method, introduced by Wampler\cite{4075580}. In it, equation (\ref{jac}) is approximately solved by 
\begin{align}
    \dot{\mathbf{q}} = \mathbf{J}(\mathbf{q})^{T} \cdot (\mathbf{J}(\mathbf{q}) \cdot \mathbf{J}(\mathbf{q})^T + \lambda^2 \cdot \mathbf{I})^{-1} \cdot \dot{\mathbf{x}},
\end{align}
where $\lambda$ is a dampening factor that serves to decrease $\dot{\mathbf{q}}$ as the robot moves into configurations near singularities. Using this, a standard inverse kinematics approach can be applied as given in algorithm \ref{alg:ik}.
\begin{algorithm}
\caption{Damped Least Squares Inverse Kinematics}\label{alg:ik}
\hspace*{\algorithmicindent} \textbf{Input}: desired workspace pose $\mathbf{x}_d$, integration step $\Delta_{\textrm{int}}$,\\ \hspace*{\algorithmicindent} \hspace*{\algorithmicindent} \hspace*{\algorithmicindent}differentiation step $\Delta_{\textrm{dif}}$, accuracy threshold $\epsilon$\\
\hspace*{\algorithmicindent} \textbf{Output}: desired configuration space pose $\mathbf{q}_d$
\begin{algorithmic}[1]
\State $\mathbf{q}_{\textrm{it}} \gets \mathbf{q}_{\textrm{c}}$
\State $\mathbf{x}_c \gets$ current workspace pose via forward kinematics from $\mathbf{q}_{\textrm{it}}$ 
\State $\mathbf{x}_e \gets \mathbf{x}_c - \mathbf{x}_d$
\Do
    \State $\dot{\mathbf{x}}_e \gets \frac{\mathbf{x}_e}{\Delta_{\textrm{dif}}}$
    \State $\dot{\mathbf{q}} \gets \mathbf{J}(\mathbf{q}_{\textrm{it}})^{T} \cdot (\mathbf{J}(\mathbf{q}_{\textrm{it}}) \cdot \mathbf{J}(\mathbf{q}_{\textrm{it}})^T + \lambda^2 \cdot \mathbf{I})^{-1} \cdot \dot{\mathbf{x}}_e$
    \State $\mathbf{q}_{\textrm{it}} \gets \mathbf{q}_{\textrm{it}} + \dot{\mathbf{q}} \cdot \Delta_{\textrm{int}}$
    \State $\mathbf{x}_c \gets$ current work space pose via forward kinematics from $\mathbf{q}_{\textrm{it}}$ 
    \State $\mathbf{x}_e \gets \mathbf{x}_c - \mathbf{x}_d$
\doWhile{$||\mathbf{x}_e|| > \epsilon$}
\State $\mathbf{q}_d \gets \mathbf{q}_{\textrm{it}}$
\State \Return $\mathbf{q}_d$
\end{algorithmic}
\end{algorithm}

\subsection{Observation Space}
\noindent The observation space consists of $O_{robot}$ and $O_{rays}$. $O_{robot}$ is a 13-dimensional vector consisting of:
\begin{itemize}
    \item 6 current joint angles in rad (for a 6-DoF robot arm).
    \item 3 scalar values containing the difference between the current position and goal position in cartesian coordinates
    \item 3 scalar values describing the current end-effector pose in Euler angles
    \item 1 scalar value for the distance from the current position to the goal position 
\end{itemize}

\begin{figure}[H]%
    \centering
    \subfloat{
    {\includegraphics[height=4cm]{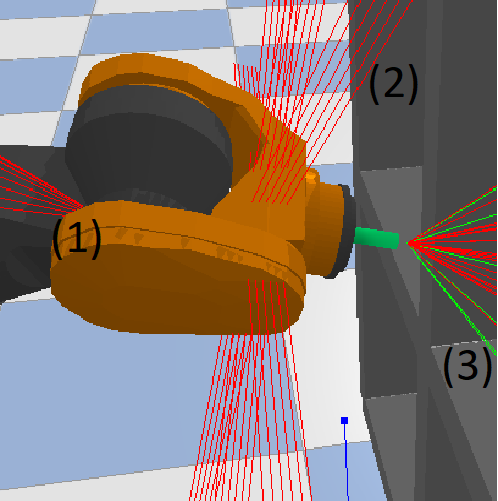} }
    }%
    \subfloat{{\includegraphics[height=4cm]{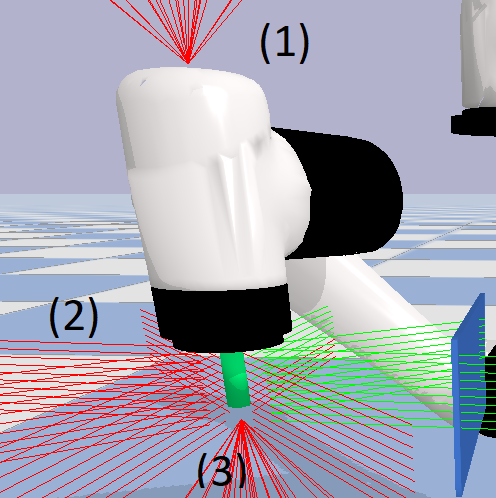} }}%
    \caption{Virtual lasers on two different robot models (left: Kuka, right: UR5). Numbers from (1)~-~(3) represent the rays from each sensor. Green rays indicate whether obstacles are detected. Each sensor has a range of 0.4m}%
    \label{fig:vlsr}%
\end{figure}

\noindent Virtual sensors are placed on the tip of the end-effector, the outer surface of the end-effector, and at the top of the wrist. Figure \ref{fig:vlsr} shows the arrangement of the rays for two different robot models. The observation $O_{rays}$ from the virtual lasers is a 129-dimensional vector consisting of the hit fractions of these 129 rays. Depending on the specific end-effector or robot model, the exact number and angle of the rays may need to be reconfigured. The observation provided by the environment to the agent is the concatenation of $O_{robot}$ and $O_{rays}$.

For both robot models we have arranged the sensors as followed:
\begin{itemize}
    \item The wrist sensor consists of 24 rays. Each ray is vertically placed by an angular distance of $\phi = 30^{\circ}$ while having a polar angle of $\vartheta = 40^{\circ} $.
    \item The end-effector’s wraparound surface is equipped with a total of 80 rays. Ten rays each are horizontally spaced by $\phi = 45^{\circ}$.
    \item 25 rays are placed on the tip of the end-effector with a polar angle of $\vartheta = 20^{\circ} $ and horizontally spaced by $\phi = 30^{\circ}$.
\end{itemize}

\subsubsection{Reward System}
\noindent The reward system needs to be carefully designed as it plays a crucial part in the behavior of the agent. Our reward function $r(s_t,a_t)$ is defined as: 

\begin{equation}
   r(s_t,a_t) = r_{suc}^t+r_c^t+r_d^t+r_{shake}^t+r_o^t+r_e^t
    \label{eq:totalreward}
\end{equation}

     \noindent $r_{suc}^t$ is the success reward. Success is considered when $d_t$ (distance to goal) is less than the distance threshold $d_{Th}$:
        \begin{equation}
        r_{suc}^t=\left\{
        \begin{aligned}
        10 & , & d_t < d_{Th} \\
        0 & , & otherwise.
        \end{aligned}
        \right.
        \label{eq:susscssreward}
        \end{equation}
     $r_c^t$ is the collision penalty:
        \begin{equation}
        r_c^t=\left\{
        \begin{aligned}
        0 & , & otherwise \\
        -10 & , & if \ robot\  collides.
        \end{aligned}
        \right.
        \end{equation}
     $r_d^t$ is the distance penalty:
        \begin{equation}
        r_d^t=-0.01\cdot d_t
        \end{equation}
        \begin{itemize}
        \item $r_o^t$ is out of bounds of workspace penalty:
        \end{itemize}
        \begin{equation}
        r_c^t=\left\{
        \begin{aligned}
        0 & , & otherwise \\
        -10 & , & if\ x_t<x_{min}\ or\ x_t>x_{max}\\
        \ & & \ or\ y_t<y_{min}\ or\ y_t>y_{max}\\
        \ & & \ or\ z_t<z{_min}\ or\ z_t>z_{max}.
        \end{aligned}
        \right.
        \label{eq:outreward}
        \end{equation}
        \begin{itemize}
        \item $r_e^t$ is the penalty for exhausting the number of steps in one episode. It prevents the robot from abandoning the exploration of space by avoiding collision:
        \end{itemize}

        \begin{equation}
        r_{e}^t=\left\{
        \begin{aligned}
        0 & , & otherwise \\
        -5 & , & step>step_{max}.
        \end{aligned}
        \right.
        \label{eq:exhreward}
        \end{equation}
        
         \noindent $r_{shake}^t$ is the penalty for unsmooth movements where $n_s$ describes the number of times the robot moved back and forth for the last 10 steps. Computation of $n_s$ can be seen in \ref{algo:shaking}. 
         \begin{equation}
        r_{shake}^t=-0.005\cdot n_s
        \label{eq:shakereward}
        \end{equation}


\begin{algorithm}[h]
  \SetAlgoLined
  \KwIn{Queue $Q_d$ for past distance recording, List $L_m$ for past direction recording}
  \KwOut{Number of uneven movements $n_{s}$}
  $n_{s}$=0 \;
  \If{$len(Q_d)>10$}
  {
    $Q_d$.popleft()
  }
  
  $Q_d$.append($d_t$)\;
  
  \For{$len(Q_d)-1$}
  {
    $L_m$.append($0$) \textbf{if} $Q_d[i+1]-Q_d[i]\ge 0$ \textbf{else} $L_m$.append($1$) 
  }
  \For{$len(L_m)-1$}
  {
    \If {$L_m[j] \ne L_m[j+1]$}
       {$n_{s} += 1$ }
  }
return $n_{s}$
  \caption{Number of uneven movements in past 10 steps}
  \label{algo:shaking}
\end{algorithm}


\subsubsection{Neural Network Architecture and Agent Design}
\begin{figure}[H]
    \centering
	\includegraphics[width=0.8\linewidth]{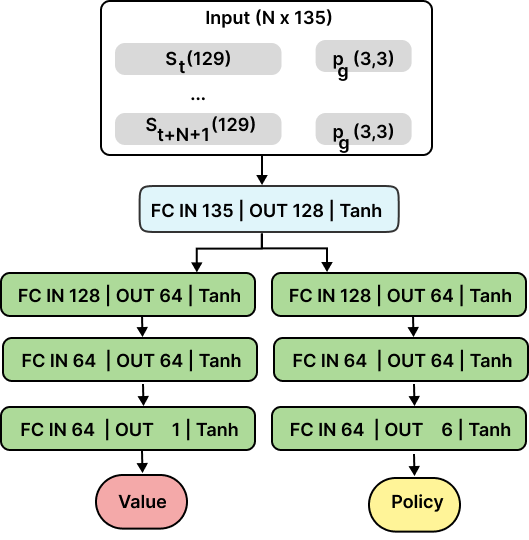}
	\caption{As input, we forward the LIDAR scan observations of N agents into the network. For both, actor- and critic networks, we use the same head.}
	\label{nn}
\end{figure}

\noindent The neural network architecture is illustrated in Fig. \ref{nn}. It consists of two networks, one for the value and one for the policy function. The input to the network is a total of 129 rays from the three virtual LIDAR scans as well as the goal position and the robot's current pose. We train on a continuous action state for more flexibility and smoothness of actions \cite{faust2018prm}. The action space $A$ is defined as follows:

  \begin{eqnarray}
    A &= \{\Delta x , \Delta y , \Delta z, \Delta\theta_y, \Delta\theta_p, \Delta\theta_r\}\quad \\
    \Delta x , \Delta y , \Delta z &\in [-0.005, 0.005]\hspace{0.5em} m \\
    \Delta\theta_y, \Delta\theta_p, \Delta\theta_r\ &\in [-0.005, 0.005]\hspace{0.5em} rad
  \end{eqnarray}

\subsubsection{Training Algorithm}

\begin{algorithm}[!h]
  \SetAlgoLined
  \KwIn{initial policy parameters $\pi_{0}$, initial value function parameters $\phi_{0}$, clipping threshold $\epsilon$}
    \For{$N1$} 
    {
        Reset the environment and get initial state $s_0$
    
        \For{$N2$} 
        {
            \For{T}
            {
                {
                    Select an action $a_t$ by running policy $\pi_{\theta_{k}}$\\
                    Execute $a_t$ in environment\\
                    Get the reward $r_t$ and next state $s_{t+1}$\\
                    Collect the transition $(s_t, a_t, r_t, s_{t+1})$ into set of trajectories $\mathfrak{D}_{k}$
                }
            }
            {
                Compute rewards-to-go $\hat{R_{t}}$ \\
                Estimate advantages $\hat{A_{t}^{\pi_{k}}}$ based on current value function $V_{\phi_{k}}$
            }
        }
        {
           maximize PPO-Clip objective:\\
            \centerline{$\theta _{k+1} =arg\max _{\theta} \mathfrak{L}_{\theta_k}^{CLIP}(\theta)$}

            Update value function:
            \centerline{$\phi_{k+1} = arg\min_{\phi}\frac{1}{\left | \mathfrak{D}_{k} \right |T }\sum_{\tau\in\mathfrak{D}_{k} }\sum_{t=0}^{T}(V_{\phi}(s_t)-\hat{R}_t)^{2} $}
        }
    }
  \caption{Proximal Policy Optimization}
  \label{alg_tr}
\end{algorithm}

\noindent We train our DRL agents using the PPO \cite{ppo} algorithm as described in \ref{alg_tr}. Internally, PPO utilizes an Advantage Actor-Critic structure, with an actor providing a policy and a critic providing a value function. Each includes its policy and value network. We chose a model in which actors and critics share a feature extractor of a certain size to extract features from inputs to save computation time. Training is parallelized by simulating multiple agents and environments simultaneously as shown in Fig. \ref{system} and \ref{nn}. We have further accelerated the training by simplifying the laser data and setting augmented targets. Beik et al. \cite{beik2019mixed} suggest an adaptive strategy for adjusting the goal's accessibility. Training a "larger-than-life" aim simplifies the effort. The augmented target size is modified based on training accomplishments. If the agent consistently fails to reach the target, the goal size is increased; otherwise, the goal size is lowered. 
\begin{equation}
\tilde{\rho}(e) =\left\{
\begin{aligned}
\rho  & , \ e<e_{\zeta } \\
\tilde{\rho}(e-1)+\delta^{+}  & ,\ \eta (k,e)<P_{\zeta},e\ge e_{\zeta },\tilde{\rho}(e)<\tilde{\rho}_{max} \\
\tilde{\rho}(e-1)-\delta^{-}  & ,\ \eta (k,e)\ge P_{\zeta},e\ge e_{\zeta },\tilde{\rho}(e)>\tilde{\rho}_{min} \\
\rho &,\ \eta (k,e)=1,e\ge e_{\zeta },\tilde{\rho}(e)=\rho\\
\tilde{\rho}(e-1)&, \qquad otherwise 
\end{aligned}
\right.
\label{eq:augmentedsize}
\end{equation}
Equation \ref{eq:augmentedsize} shows the definition of our augmented target size $\tilde{\rho}(e)$. Here, $\tilde{\rho}(e)$ is the augmented target size at episode $e$, $e_{\zeta }$ is the episode number in which the size remains unchanged. Target size increment and decrement values are represented by $\delta^{+}$ and $\delta^{-}$ respectively. The success threshold is described by $P_{\zeta}$ while $\eta (k,e)$ indicates the number of successes in reaching the (augmented) target in the past $k$ episodes.
\begin{equation}
\eta(k,e)=\frac{1}{k} \sum_{e'=e-k}^{e} S_{e'}
\label{eq:sucinkepi}
\end{equation}
$S_e$ is the success/failure of the agent at the end of the episode $e$:
\begin{equation}
S_e =\left\{
\begin{aligned}
1  & , & if\ ||P_{t}-P_{target}|| \le \tilde{\rho}(e) \\
0  & , & otherwise
\end{aligned}
\right.
\label{eq:success}
\end{equation}
The parameters used for the augmented target size can be found in Table \ref{tab:augm}.

\subsubsection{Training Setup}
\noindent The agent is trained in randomized environments. Static and dynamic obstacles are spawned randomly after each episode. The amount of obstacles increases/decreases depending on the agent's success rate. The training was optimized with GPU usage and trained on an NVIDIA RTX 3090 GPU. Training time took around 6h to converge. 
\begin{figure}[h!]%
    \centering
    \subfloat{{\includegraphics[width=0.5\linewidth]{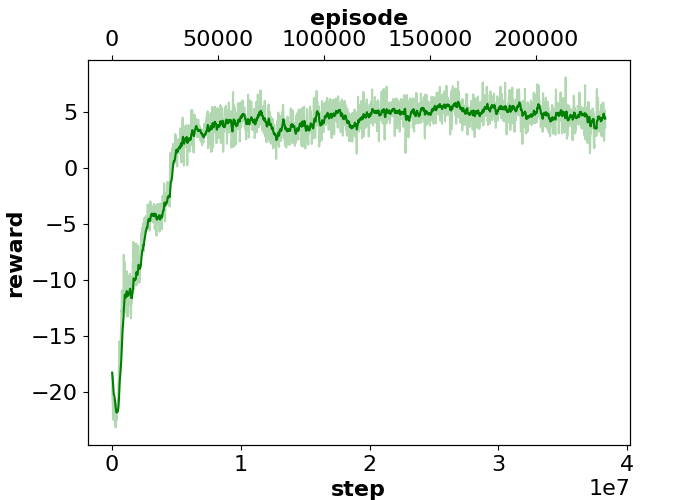} }}%
    \subfloat{{\includegraphics[width=0.5\linewidth]{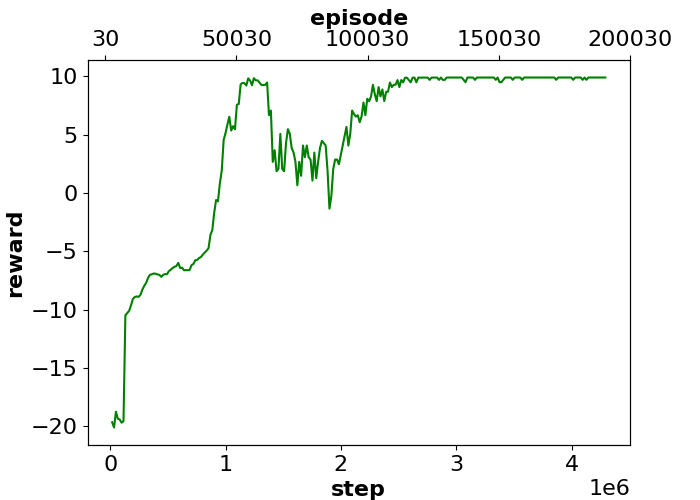} }}%
    \caption{Learning curve of the training process for both robots (left: UR5, right: Kuka).}%
    \label{fig:train}%
\end{figure}
\noindent Figure \ref{fig:train} shows the training process for both robot models. The hyperparameters are listed in Table \ref{tab:hyper}.

%% file: content/4-evaluations.tex
\begin{figure*}[!htp]
	\centering
	\includegraphics[width=0.9\textwidth]{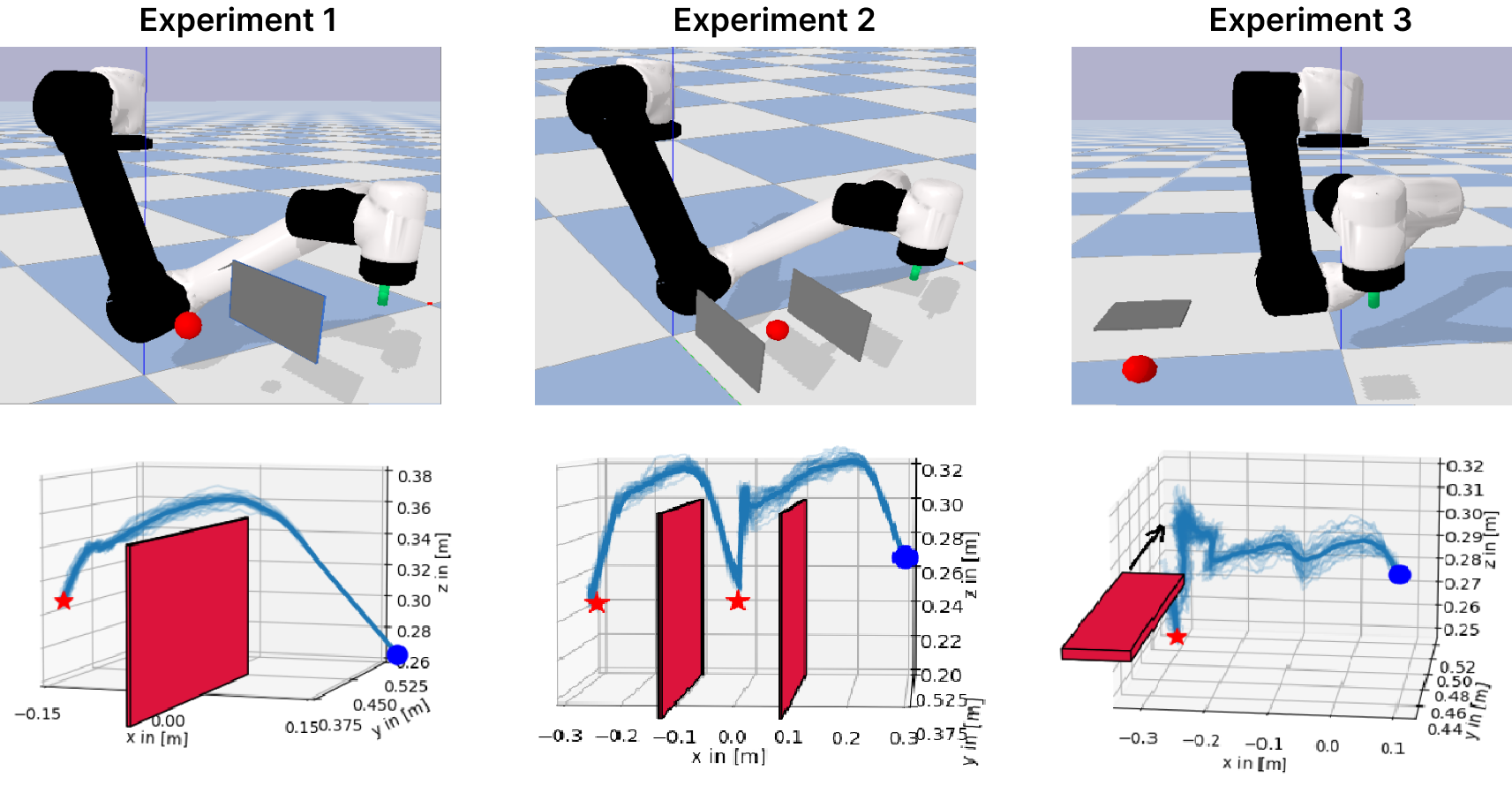}
\caption{Initial and goal positions are visualized as blue circles and red stars, respectively. Bold trajectories represent the average path of all individual runs. Experiment 3 includes a moving obstacle (arrow indicates the direction of movement). Only our DRL planner is illustrated since the sampling-based planners show high variations and occupy too much space due to their much larger trajectories.}\label{ur5:qual}
\end{figure*}

\begin{figure*}[!htp]
    \centering
    \includegraphics[width=0.7\textwidth]{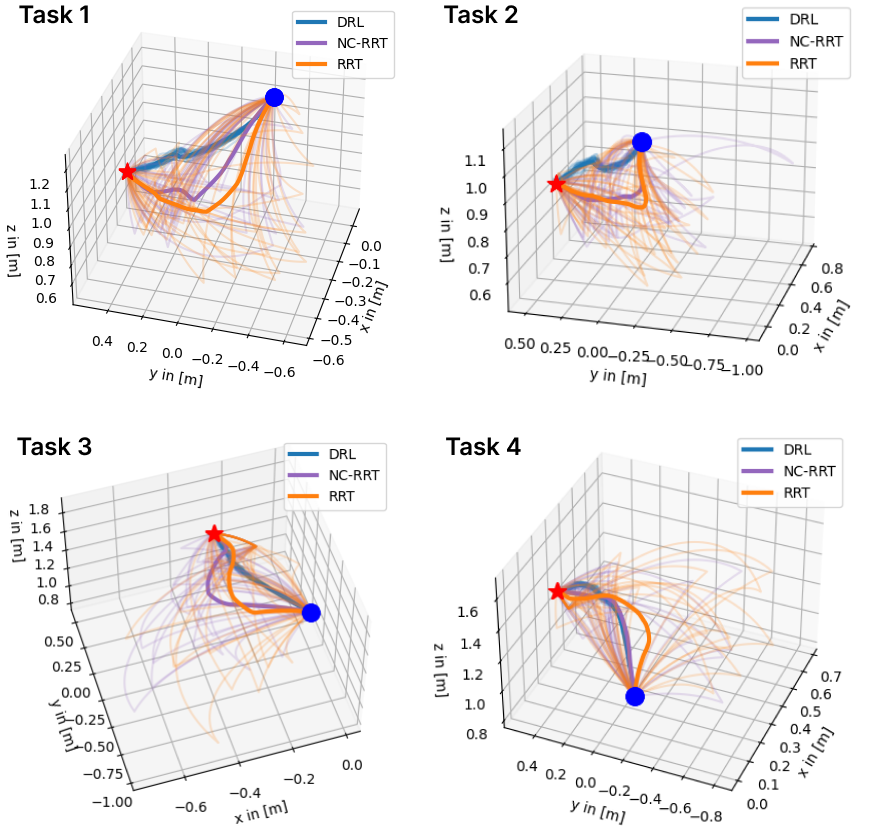}
    \caption{Trajectories of the Kuka experiments}\label{kuka_exp}
\end{figure*}

\begin{figure*}[!htp]
\centering
\includegraphics[width=0.9\textwidth]{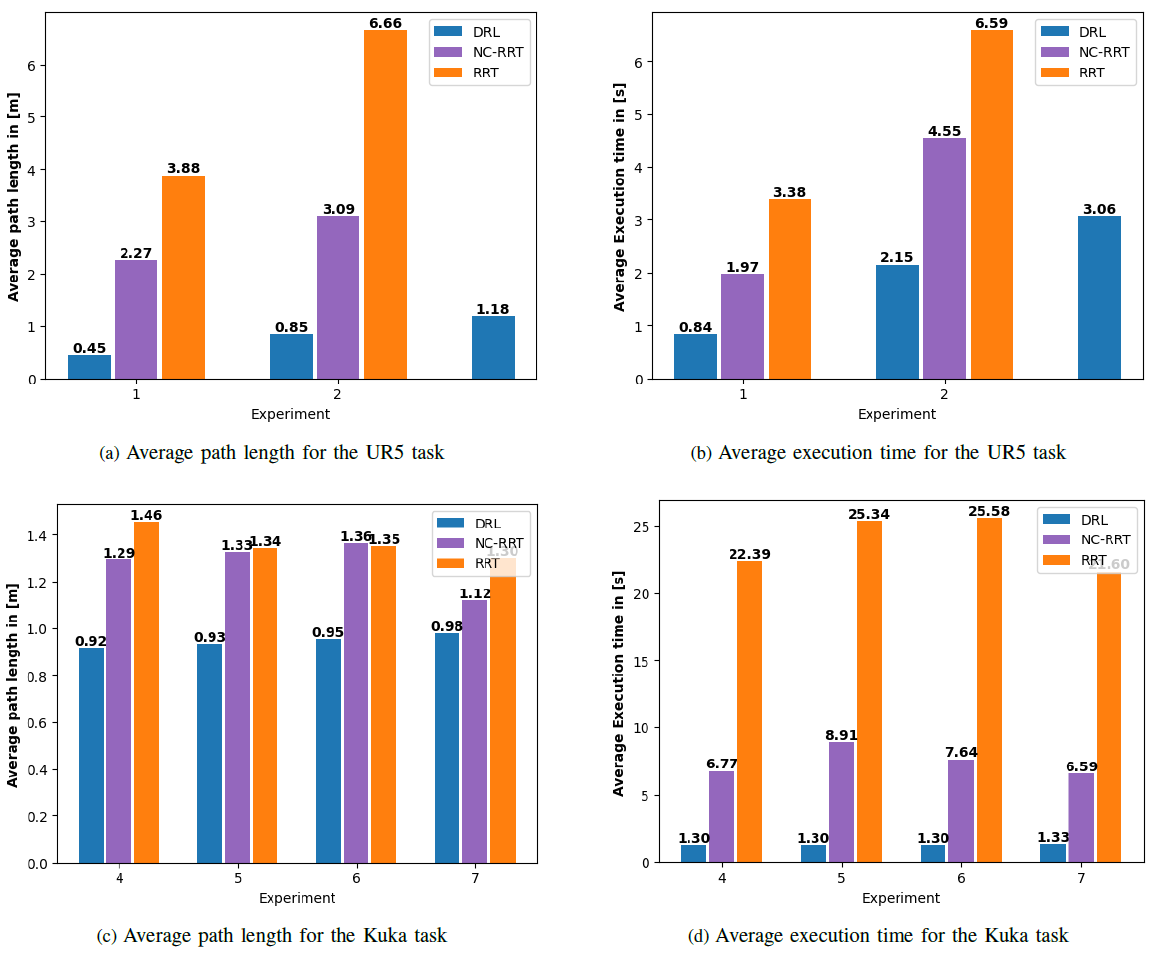}
\caption{Quantitative results of all experiments\label{eval:quant}}
\end{figure*}

\noindent In the following chapter, we will present our experiments, the results, and evaluations. This includes a comparison of two sampling-based planners RRT and a novel version of RRT called NC-RRT \cite{rrt7b} with our proposed DRL planner. We conduct experiments on two different robots (UR5 and Kuka KR 16) in which we created a total of seven different test scenarios of increasing difficulty. For each planner, we conduct 30 test runs on the same scenario. 
Figure \ref{ur5:qual} shows three experiments with different types of obstacles in form of metal plates. The UR5 is forced to avoid these plates while moving to its goal. Below each experiment, the corresponding trajectories are illustrated.   



\begin{figure}[H]%
    \centering
    \subfloat{{\includegraphics[height=4.5cm]{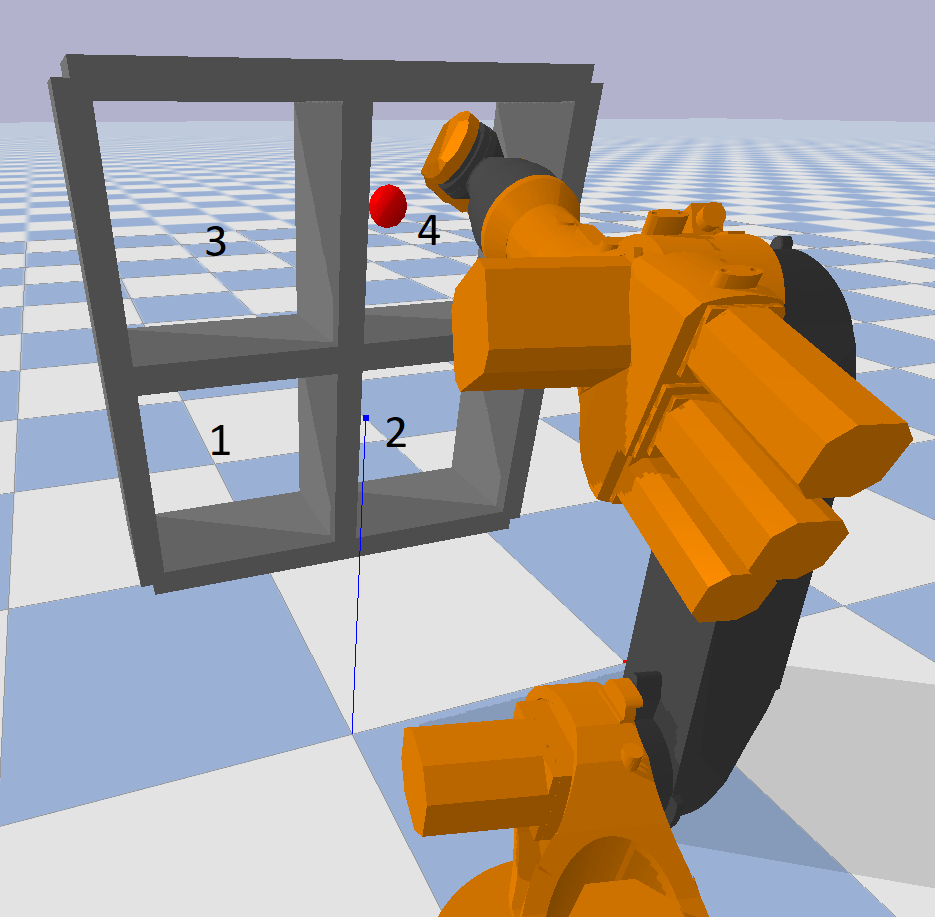} }}%
    \subfloat{{\includegraphics[height=4.5cm]{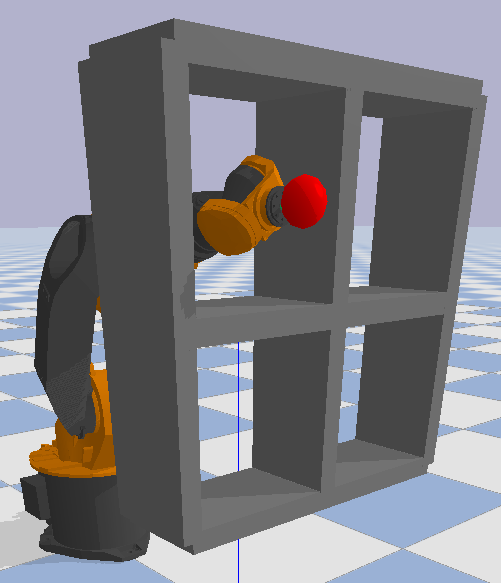} }}%
    \caption{The Left and right illustrations show the same experiment from different angles}%
    \label{kuka_sim}%
\end{figure}

\noindent While the UR5 experiments are meant to show the general path-planning capabilities of our model, the Kuka experiments will focus on industrial applications. In tasks such as welding, painting, or assembly the robot arm needs to reach certain locations in complex and narrow environments. Figure \ref{kuka_sim} shows the simulation environment of the experiment, consisting of the robot (Kuka KR 16) and a shelf-like structure representing the obstacle. The task is to reach through each space enumerated from 1 to 4. 

\noindent Figure \ref{kuka_exp} shows the qualitative results of the Kuka experiments. The initial position of the end effector is marked by a blue sphere while the goal position is indicated by a red star. Bold lines represent the average trajectory of the respective planner.

\section{Discussion}

\subsection{UR5 Trajectories}
\noindent Fig. \ref{ur5:qual} shows three simulated experiments and their respective qualitative results for the DRL planner. The dsmpling-based planners are not included as their trajectories indicate significant deviations on each run. Due to the chaotic and unstructured nature of their trajectories, a comparison would not be useful. Experiment 1 includes a single static obstacle representing a metal plate. The trajectories show how the robot's end effector is raised in order to avoid the plate. In experiment 2, an additional goal and plate are added forcing the robot to avoid both obstacles subsequently. Experiment 3 is designed to test the responsiveness of our model and contains a moving obstacle (v=0.2m/s) set to block the goal position from the end effector. Therefore, the agent raised the end effector near the goal position reaching it after the obstacle moves away. In each experiment, the trajectories are similar on each run indicating a high consistency of the DRL planner. 

\subsection{Kuka Trajectories}
\noindent The qualitative trajectories of all planners on each task are illustrated in Fig. \ref{kuka_exp}. It is noticeable that the trajectories of the sampling-based planners show strong deviations from run to run. Meanwhile, the trajectories of our DRL model seem to be more consistent as the individual trajectories are almost indistinguishable from the average trajectory. Furthermore, trajectories of the sampling-based planners indicate much larger paths.

\subsection{Quantitative Evaluations}


\noindent Figure \ref{eval:quant} shows the quantitative evaluations of our experiments for all three planners. In total, we conducted 30 test runs for each planner on 7 different experiments. We evaluate the efficiency of path planning by calculating the average distance traveled and the average time to reach the goal. All planners concluded the experiments without any collision.

\subsubsection{UR5}
\noindent Fig. \ref{eval:quant} (a) shows the average trajectories for experiments 1-3. The DRL model produces significantly shorter trajectories compared to the sampling-based planners. Shorter trajectories translate to faster execution times as shown in Fig. \ref{eval:quant} (b). It should be noted that the high execution times of the sampling-based planners are not solely caused by larger trajectories but also by high computation time. In experiment 1, our DRL planner finishes almost twice as fast as NC-RRT and four times faster than RRT. Experiment 2 shows similar results as the DRL planner is again twice as fast as NC-RRT and three times faster than RRT. For experiment 3, the classical approaches were unable to cope with the dynamic obstacle due to their long planning times and failed altogether. Thus, only the DRL approach could be evaluated.

\subsubsection{Kuka}
\noindent The shelf obstacle is the most complex environment in all the experiments as it contains multiple narrow tunnels, which increase the possibility of collisions. Similar to experiments 1 and 2, the DRL approach outperforms the RRT planners in all instances. Fig. \ref{eval:quant} (c) and (d) signify the consistency of the DRL planner as path length and time are mostly the same for experiments 4-7, while the RRT planner shows minor discrepancies. Particularly the DRL planner is about six times faster than NC-RRT and up to 19 times faster than RRT.

%% file: content/5-conclusion.tex
\section{Conclusion}

\noindent In this paper, we proposed a DRL-based motion planning model
for industrial robots. Results show that our model significantly outperforms both sampling-based planners in all experiments.
While in simple environments (experiments 1-2), our model is almost twice as fast as NC-RRT in more complex environments (experiments 4-7) NC-RRT is outperformed by almost six times. It can be concluded that in contrast to sampling-based methods, our proposed planner is less sensitive to environmental complexity. Furthermore, our planner is able to react to unexpected situations, for example, if an object is mistakenly misplaced due to its use of laser sensors.
In future works, we aspire to extend our model by incorporating additional functionalities besides path planning. For example, in welding tasks, the robot not only has to move to the target location but also assume the desired pose with its end effector (torch). Traditionally, each welding spot needs to be handcrafted by skilled workers or third-party tools. We will try to develop an end-to-end model for path planning and pose estimation. Additionally, we will further improve our model by including additional types of sensors such as RGB cameras. In collaborative environments, the use of sensor-based planners is crucial as the robot has to be able to perform tasks while avoiding dynamic obstacles such as humans or other robots. Another challenge that still remains is the sim-to-real gap. While in the simulation we showed promising results with the use of virtual sensors, the approaches have to be verified on real robots. However, in offline programming, our DRL planner may be a more efficient alternative to classical motion planning methods.

\appendix
\noindent Code available at: https://github.com/ignc-research/IR-DRL
\newpage
\begin{table}[ht]
\begin{tabular}{|l|l|}
\hline
\rowcolor[HTML]{EFEFEF} 
\textbf{Hyperparameter}                                                      & \textbf{value} \\ \hline
Discount factor                                                              & 0.99           \\ 
Batch size                                                                   & 256            \\ 
Learning rate                                                                & 0.0003         \\ 
Clip-range                                                                   & 0.2            \\ 
Max steps in one episode                                                     & 1024           \\ 
Entropy coefficient for the loss calculation                                 & 0.0            \\ 
Value function coefficient for the loss calculation                          & 0.5            \\ 
Max Gradient                                                                 & 0.5            \\ 
Trade-off factor                                                             & 0.95           \\ \hline
\end{tabular}%
\caption{Hyperparameter for training}
\label{tab:hyper}
\end{table}
\begin{table}[ht]
\begin{tabular}{|c|c|c|c|c|c|c|}
\hline
\rowcolor[HTML]{EFEFEF} 
$\tilde{\rho}_{max}$ & $\tilde{\rho}_{min}$ & $\delta^{+}$ & $\delta^{-}$ & $e_{\zeta }$ & $P_{\zeta}$ & $k$  \\ \hline
0.1  & 0.01 & 0.001 & 0.01 & 1e3 & 0.9 & 50 \\ \hline
\end{tabular}%
\caption{Parameters for augmented target size}
\label{tab:augm}
\end{table}

\begin{table}[H]
\begin{tabular}{c|c|c|c|c|c|c|}
\cline{2-7}
                                          & \multicolumn{3}{c|}{\textbf{execution time in {[}s{]}}}               & \multicolumn{3}{c|}{\textbf{path length in {[}m{]}}}                                       \\ \cline{2-7} 
                                          & \multicolumn{6}{c|}{Planner}                                                                                                                                       \\ \hline
\multicolumn{1}{|c|}{\textbf{Experiment}} & DRL                          & RRT                           & NC-RRT & DRL                          & RRT                          & NC-RRT                       \\ \hline
\multicolumn{1}{|c|}{1}                   & \cellcolor[HTML]{EFEFEF}0.84 & \cellcolor[HTML]{FFCCC9}3.38  & 1.97   & \cellcolor[HTML]{EFEFEF}0.45 & \cellcolor[HTML]{FFCCC9}3.88 & 2.27                         \\ \hline
\multicolumn{1}{|c|}{2}                   & \cellcolor[HTML]{EFEFEF}2.15 & \cellcolor[HTML]{FFCCC9}6.59  & 4.54   & \cellcolor[HTML]{EFEFEF}0.85 & \cellcolor[HTML]{FFCCC9}6.65 & 3.08                         \\ \hline
\multicolumn{1}{|c|}{3}                   & \cellcolor[HTML]{EFEFEF}3.05 & /                             & /      & \cellcolor[HTML]{EFEFEF}1.18 & /                            & /                            \\ \hline
\multicolumn{1}{|c|}{4}                   & \cellcolor[HTML]{EFEFEF}1.29 & \cellcolor[HTML]{FFCCC9}22.38 & 6.76   & \cellcolor[HTML]{EFEFEF}0.91 & \cellcolor[HTML]{FFCCC9}1.45 & 1.29                         \\ \hline
\multicolumn{1}{|c|}{5}                   & \cellcolor[HTML]{EFEFEF}1.29 & \cellcolor[HTML]{FFCCC9}25.34 & 6.36   & \cellcolor[HTML]{EFEFEF}0.93 & \cellcolor[HTML]{FFCCC9}1.34 & 1.32                         \\ \hline
\multicolumn{1}{|c|}{6}                   & \cellcolor[HTML]{EFEFEF}1.30 & \cellcolor[HTML]{FFCCC9}25.58 & 5.46   & \cellcolor[HTML]{EFEFEF}0.95 & 1.35                         & \cellcolor[HTML]{FFCCC9}1.36 \\ \hline
\multicolumn{1}{|c|}{7}                   & \cellcolor[HTML]{EFEFEF}1.33 & \cellcolor[HTML]{FFCCC9}21.59 & 4.70   & \cellcolor[HTML]{EFEFEF}0.98 & \cellcolor[HTML]{FFCCC9}1.30 & 1.12                         \\ \hline
\end{tabular}%
\caption{Mean values from 30 runs for all experiments. Best and worst performances are colored grey and red respectively.}
\label{tab:eval}
\end{table}